\documentclass{article}
\usepackage{spconf,amsmath,epsfig}

\usepackage{graphicx} 
\usepackage{subfigure} 

\usepackage{amssymb}

\usepackage{algorithm}
\usepackage{algorithmic}

\begin{document}\sloppy

\def\x{{\mathbf x}}
\def\L{{\cal L}}

\title{ADAPTIVE AFFINITY MATRIX FOR UNSUPERVISED METRIC LEARNING}
%
\name{Yaoyi Li, Junxuan Chen, Yiru Zhao, Hongtao Lu\footnotemark{*}\thanks{$^*$Corresponding author}\thanks{This paper is supported by NSFC (No. 61272247, 61533012, 61472075), the 863 National High Technology Research and Development Program of China (SS2015AA020501) and the Major Basic Research Program of Shanghai Science and Technology Committee (15JC1400103).}}
\address{Key Laboratory of Shanghai Education Commission for Intelligent Interaction and Cognitive Engineering\\
	Department of Computer Science and Engineering, Shanghai Jiao Tong University, P.R.China\\
	\{dsamuel, chenjunxuan, yiru.zhao, htlu\}@sjtu.edu.cn}
%
%
%

\maketitle

\begin{abstract}
	Spectral clustering is one of the most popular clustering approaches with the capability to handle some challenging clustering problems. Only a little work of spectral clustering focuses on the explicit linear map which can be viewed as the distance metric learning. In practice, the selection of the affinity matrix exhibits a tremendous impact on the unsupervised learning. In this paper, we propose a novel method, dubbed Adaptive Affinity Matrix (AdaAM), to learn an adaptive affinity matrix and derive a distance metric. We assume the affinity matrix to be positive semidefinite with ability to quantify the pairwise dissimilarity. Our method is based on posing the optimization of objective function as a spectral decomposition problem. The provided matrix can be regarded as the optimal representation of pairwise relationship on the manifold. Extensive experiments on a number of image data sets show the effectiveness and efficiency of AdaAM. 
\end{abstract}


\section{Introduction}
\label{Introduction}


Spectral clustering methods which are based on eigendecomposition demonstrate splendid performance on many real-world challenge data sets. During the past decades, a series of spectral clustering methods have been proposed: Multidimensional Scaling (MDS) \cite{cox2000multidimensional}, Local Linear Embedding (LLE)  \cite{roweis2000nonlinear}, Isomap  \cite{tenenbaum2000global}, Laplacian Eigenmaps  \cite{belkin2001laplacian} and variant of Spectral Clustering  \cite{ng2002spectral}. 
There are three shortages of spectral clustering methods mentioned above. First, these approaches only provide the embedding map of the training data. The out-of-sample extension is not straightforward. Second, The complexity of these approaches relies on the number of data points. 
Third, the performance of spectral clustering methods highly depend on the robustness of the affinity graph. 

Many important progresses  \cite{bengio2004out,niyogi2004locality,fowlkes2004spectral,yan2009fast,chen2011large,pavan2007dominant,premachandran2013consensus,zhu2014constructing,nie2014clustering} have been made to mitigate the above issues of the spectral clustering. Locality Preserving Projections (LPP) proposed in \cite{niyogi2004locality} introduces a linear projection obtained from Laplacian Eigenmaps. Their work provides a linear approximation of the embedding mapping, which reduces the time complexity and achieves out-of-sample extension straightforwardly. The linear embedding gives a metric learning perspective of the spectral clustering. 
Nie, Wang, and Huang proposed the Projected Clustering with Adaptive Neighbors (PCAN)  in \cite{nie2014clustering} where they regard the pairwise similarity as an extra variable to be solved in the optimization problem and they set a penalty of the rank of graph Laplacian to restrict specific connected components in the affinity matrix. With this framework, PCAN alternately update affinity matrix and projection.
Although some affinity learning algorithms have been proposed in recent years, the technique of choosing an appropriate affinity matrix is still remained to be addressed.

Our goal is to extract more adaptive similarity information with minimal extra time consumption for the linear approximation of spectral clustering. Such information will take the objective of locality preserving rather than only the distance between images into consideration. Inspired by the recent progress on scalable spectral clustering  \cite{chen2011large} and data similarity learning  \cite{nie2014clustering}, we propose a novel approach dubbed Adaptive Affinity Matrix (AdaAM). Our affinity matrix is relatively dense and can capture both global and local information. Specifically, AdaAM decomposes the affinity graph into a product of two low-rank identical matrices. As the ideal case described in \cite{ng2002spectral}, if we assume the pairwise affinity in the same class are exceedingly similar, the affinity matrix may turn into a low-rank matrix. We optimize the decomposed matrix with the similar scheme of spectral clustering. The affinity graph obtained by optimization is used as an 
intermediate affinity matrix, firstly. 
With the combination of the 
intermediate affinity matrix and the $k$-NN affinity graph derived by the heat kernel, we figure out a final adaptive affinity matrix
from a naive spectral clustering. We conduct the affinity graph with the data projection and apply LPP to this specific graph  to learn a metric for clustering.

%
%

We illustrate the effective and efficiency of the proposed approach for clustering on image data sets. We show the advantage of AdaAM 
for challenging data sets by comparing our approach with $k$ nearest neighborhood heat kernel ($k$NN)  \cite{belkin2001laplacian} and some other state-of-the-art algorithms in Section \ref{secExp}. 

Our main contribution is that we integrate the affinity matrix learning into the framework of spectral clustering with the same paradigm, and we employ the low rank trick to make our approach more efficient.

\section{Adaptive Affinity Matrix}

\subsection{Notation}
In this paper, we write all matrices as uppercase (English or Greek alphabet) and vectors are written as lowercase. 
The vector with all elements one is denoted by $\bf{1}$.
$H$ is the  centering matrix denoted by $H = I-\frac{1}{n}\bf{1}\bf{1}^T$. The origin data matrix is denoted by $X \in \mathbb{R}^{n\times d}$, where $n$ is the number of the data points and $d$ is the dimension of the data. $X$ is assumed to be normalized with zero mean, i.e. $X = HX$. The denotation $x_i$ means the $i$-th data point vector. We also denote the linear projection by $A$ and denote the metric matrix by $M = A^TA$. 
Hence the Mahalanobis distance based on is $dis_m(x_i, x_j) = (x_i - x_j)^TM(x_i - x_j)$. 
The $k$-NN heat kernel matrix is denoted by $W \in \mathbb{R}^{n \times n}$ with 
\begin{equation}
	w_{ij} = \begin{cases} \mathrm{exp}(-\frac{\|x_i-x_j\|_{2}}{t}), \; &x_i\in\mathcal{N}_k(x_j)\;\mathrm{or}\; x_j\in\mathcal{N}_k(x_i)\\
		0,&\mathrm{otherwise}\end{cases}                                     
\end{equation}
where $\mathcal{N}_k(x)$ is the set of $k$ nearest neighbors of $x$. The corresponding Laplacian matrix is denoted by $L=D-W$, where $D$ is the diagonal matrix with $d_{ii} = \sum_j w_{ij}$. We also denote both intermediate increment and final adaptive affinity matrix as $\Delta$, the corresponding diagonal weight matrix and Laplacian matrix as $D_\Delta$ and $L_\Delta = D_\Delta-\Delta$.

\subsection{Intermediate Affinity Matrix}
\label{intermediate}
We separate our algorithm into two parts, intermediate affinity matrix and final adaptive affinity matrix. In this section, we will introduce the first part. For the $i$-th data point $x_i$, we connect any the data point $x_i$ to the data point $x_j$ with the similarity $\delta_{ij}$. With the hope that small Euclidean distance between two data points leads to a large similarity, we aim to choose $\delta_{ij}$ to minimize the following objective function
\begin{equation}
	\mathop{\mathrm{min}} \sum_{i,j}^{n} \|x_i-x_j\|_2^2\;\delta_{ij}
\end{equation}
under appropriate constraints, where $\delta_{ij}$ is the $ij$-th element of the intermediate affinity matrix $\Delta$. 

Different from PCAN \cite{nie2014clustering}, we reformulate the equation with graph Laplacian,
\begin{equation}
	\mathop{\mathrm{min}}\; tr(X^TL_\Delta X)
	\label{eqDelta}
\end{equation}
under some constraints.

With a straightforward thought we can decompose the Laplacian into two identical matrices, since the graph Laplacian is a positive semidefinite matrix in general. We show this thought is not appropriate in our framework as follows.

If we assume that
\begin{equation}
	L_\Delta = UU^T
\end{equation}
where $U\in \mathbb{R}^{n\times s}$ is a column orthogonal matrix with $U^TU = I$. With the relaxing of the constraints, 
we finally need to solve the 
problem
\begin{equation}
	\begin{split}
		&U = \mathop{\mathrm{arg\;min}}_{U^TU=I}\; tr(X^TUU^TX) \\
		&\Rightarrow U = \mathop{\mathrm{arg\;min}}_{U^TU=I} \;tr(U^TXX^TU)
	\end{split}
	\label{simpEigmap}
\end{equation}

If we assume the product of matrix $X$ to be $K$ (i.e. $K = XX^T$), the Eq. (\ref{simpEigmap}) gives a simple form of the Laplacian Eigenmaps

This optimization problem can be solved by selecting eigenvectors of matrix $K$ corresponding to several smallest eigenvalues.  However, $K$ is a low-rank matrix generally with  $d\ll n$ and the eigenvectors of $K$ minimizing the objective function in Eq. (\ref{simpEigmap}) is in the null space of $X$. Hence, the solution of above problem is not unique. Inspired by LSC \cite{chen2011large} we assume the affinity matrix to be a positive semidefinite matrix and decompose it into the product of a matrix $P\in \mathbb{R}^{n\times t}$ with orthogonal columns and $P^T$ instead of decomposing the  Laplacian matrix, where $t$ is the expected rank of $\Delta$.

Therefore we reformulate Eq. (\ref{eqDelta}) as
\begin{equation}
	\mathop{\mathrm{min}}_{P^TP=I} tr(X^TD_\Delta X)+tr(X^T(-PP^T)X)
	\label{eqXLX}
\end{equation}
where we abandon the properties that connected weight is non-negative and the graph Laplacian is positive semidefinite. The negative connected weights in $\Delta$ can be used to measure the dissimilarity between data points. We will show that the solution of this optimization problem makes $D_\Delta$ equal to $\bf0$.

For the first part of Eq. (\ref{eqXLX}), we can write it as
\begin{equation}
	\begin{split}
		\mathop{\mathrm{min}}\;&\sum_{i=1}^{n} \|x_i\|_2^2\;d_{\Delta ii}\\
		s.t. \;\; &P^TP=I\\
		&d_{\Delta ii}=(PP^T\textbf{1})_i
	\end{split}
	\label{eqXD}
\end{equation}

Let $z=(\|x_1\|_2^2, \|x_2\|_2^2, ... , \|x_n\|_2^2)^T$. With a Lagrange multipliers $\lambda$, the one dimensional situation of problem (\ref{eqXD}) can be rewritten as
\begin{equation}
	\mathop{\mathrm{min}}\; z^Tpp^T\textbf{1}-\lambda (p^Tp-1)
\end{equation}

Finally, the minimization problem (\ref{eqXD}) reduces to finding the eigenvector corresponding to the minimum eigenvalue of the problem $\textbf{1}z^Tp=\lambda p$. 
Because the matrix $\textbf{1}z^T$ has rank one, there is only one nonzero eigenvalue $\sum_{i=1}^{n}\|x_i\|_2^2$, which implies $\lambda = 0$.
Hence, for the $P$ satisfying problem (\ref{eqXD}) with arbitrary column number less than $n$, we have $z^TPP^T\textbf{1}=0$. It is equivalent to 
\begin{equation}
	\sum_{i=1}^{n} \|x_i\|_2^2\;d_{\Delta ii} = 0
\end{equation}

Generally, in real-world data set, $\|x_i\|_3^2\neq0$ always holds, thus, the $P$ minimizing the first part of the objective function (\ref{eqXLX}) has the property $D_\Delta = \textbf{0}$. Meanwhile the set of all $P$ with the property $D_\Delta = 0$ is the solution of Eq. (\ref{eqXD}).  

\begin{figure}[t]
	\centering
	\subfigure[]{
		\centering
		\includegraphics[width=0.45\columnwidth]{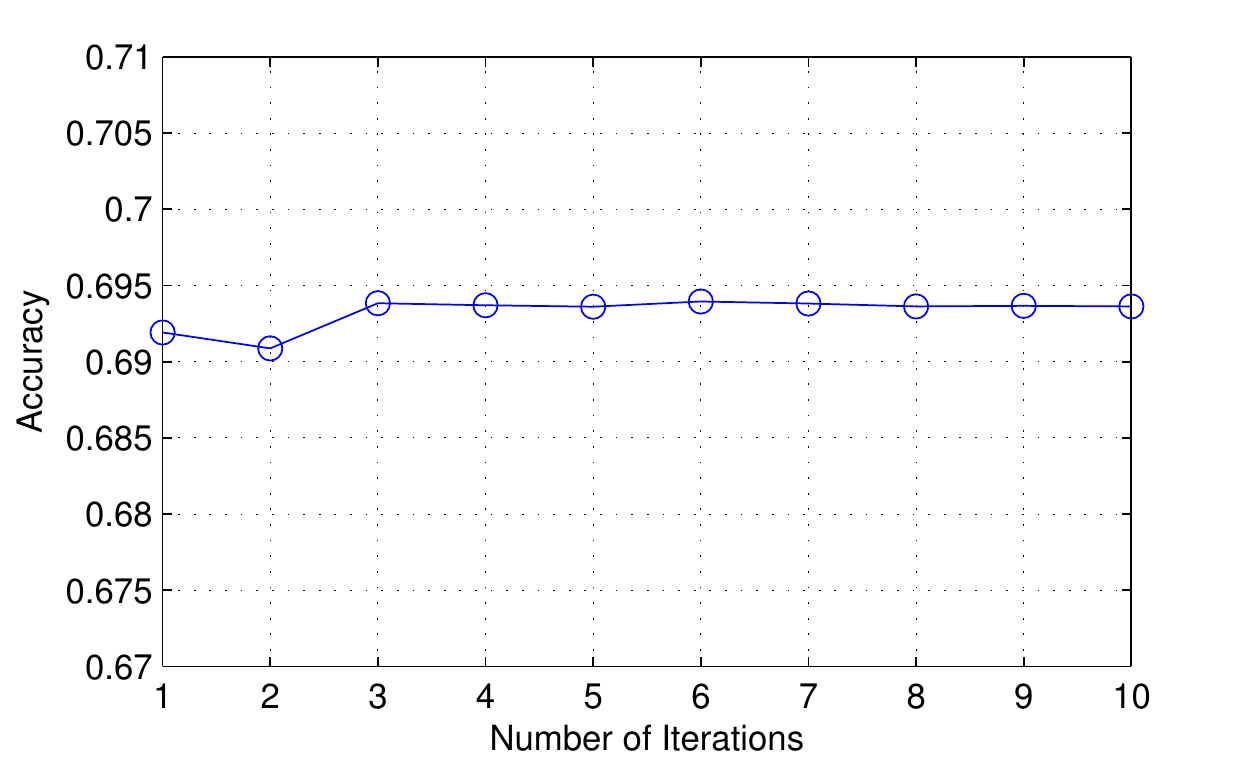}
		\label{figitera}
	}
	\subfigure[]{
		\centering
		\includegraphics[width=0.45\columnwidth]{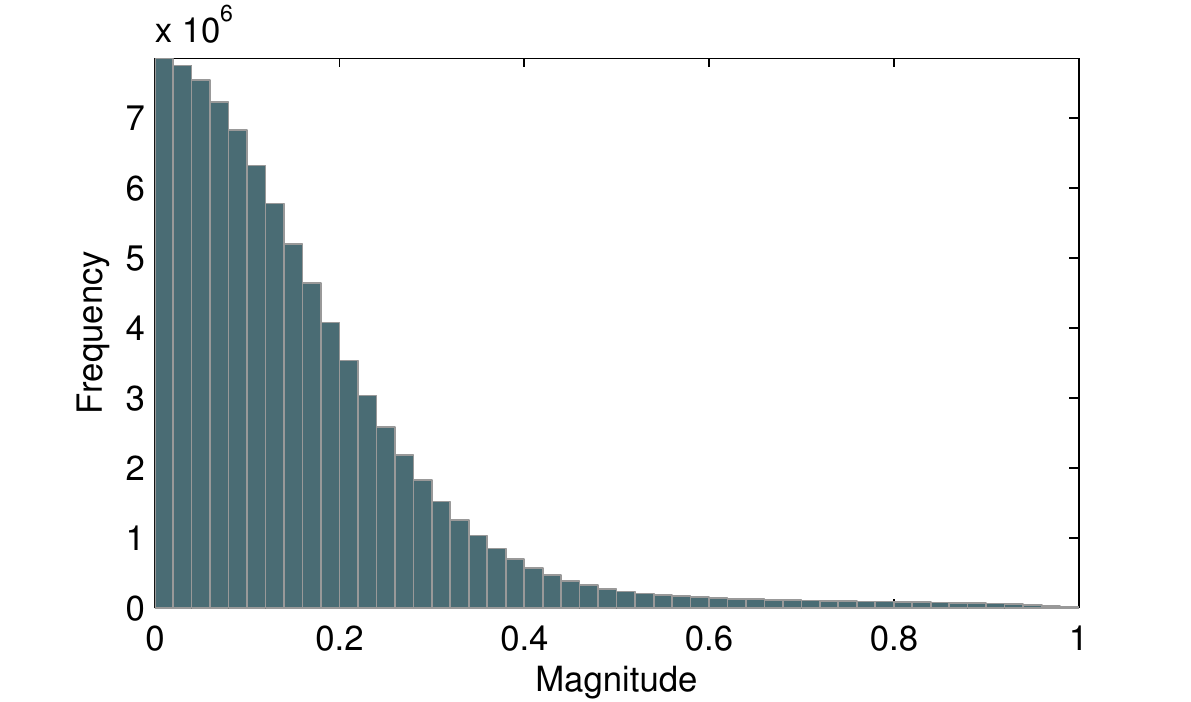}
	}
	\caption{(a)The evaluation of the clustering performance with different times iterative computation on the data set USPS. The contribution to accuracy made by iteration is less than 0.5\%. (b)The histogram of the element magnitude of the final adaptive affinity matrix obtained from data set USPS. }
	\label{figHist}
\end{figure}

The matrix $P$, which minimizes the second part of the objective function (\ref{eqXLX}), is given by the maximum eigenvalue to the eigen problem:
\begin{equation}
	\begin{split}
		(XX^T)p = \lambda p
		\Rightarrow {\bf1}XX^Tp = \lambda {\bf1}p
	\end{split}
	\label{eqSolu}
\end{equation}

As the data $X$ has zero mean, we have $\lambda {\bf1}p = {\bf1}XX^Tp = 0$. Therefore, for the maximum eigenvalue which is larger than $0$, the corresponding eigenvector always satisfies ${\bf1}p = 0$. Let the minimum solution of the second part of problem (\ref{eqXLX}) be $P=(p_1,p_2,...,p_t)$. We have  
\begin{equation}
	{\bf1}^T P = {\bf0} \Rightarrow {\bf1}^T PP^T = {\bf0} \Rightarrow d_{\Delta ii} = 0 
\end{equation}
which means that the property $D_\Delta = \textbf{0}$ holds for the optimal solution of the second part of (\ref{eqXLX}) and the solution is in the set of the solution of Eq. (\ref{eqXD}). 
Therefore the solution of the second part of Eq. (\ref{eqXLX}) can also optimize the object function (\ref{eqXD}) and the solution of the optimization problem (\ref{eqXLX}) makes $D_\Delta$ equal to $\bf0$. The objective function (\ref{eqXLX}) can be reduced to
\begin{equation}
	P=\mathop{\mathrm{arg\;max}}_{P^TP=I} \;tr(P^TXX^TP)
	\label{eqPXXP}
\end{equation}
which has the solution as singular value decomposition of $X$ with complexity relies on $d$ rather than $n$. We obtain the intermediate affinity matrix $\Delta = PP^T$ from the distribution of origin data with similarity and dissimilarity information. The graph Laplacian of $\Delta$ is $L_\Delta = D_\Delta-\Delta =-\Delta $. 

To mitigate the impact of noise and rank reducing problem, we apply sparsification to $\Delta$. 
We will discuss the sparsification further in Section \ref{sparse}.

\subsection{Final Adaptive Affinity Matrix}
\label{final}
In this section, we formulate a naive linear spectral clustering and provide the final adaptive affinity matrix.

With the intermediate affinity matrix $\Delta$, we can solve the following problem for a linear projection $A$:
\begin{equation}
	a = \mathop{\mathrm{arg\;min}}_{a^Ta=1}\; tr(a^TX^T(L+L_\Delta)Xa)
	\label{eqAXLXA}
\end{equation}
where $a$ is the one-dimension case of $A$ and $L+L_\Delta$ is the combination of the Laplacian of $k$-NN heat kernel and the intermediate affinity matrix. The projection vector $a$ is given by the minimum eigenvalue of the eigen problem:
\begin{equation}
	X^T(L-\Delta)Xa = \lambda a
\end{equation}

\begin{algorithm}[t]
	\renewcommand{\algorithmicrequire}{\textbf{Input:}}  
	\renewcommand\algorithmicensure {\textbf{Output:} }  
	\caption{Adaptive Affinity Matrix}
	\label{alg1}
	\begin{algorithmic}[1]
		\REQUIRE ~~\\
		Data points $X \in \mathbb{R}^{n \times d}$; cluster number $c$; neighborhood size $k$; reduced dimension $m$;
		\ENSURE ~~\\
		Mahalanobis metric $M$ and linear projection $A$.
		\STATE Construct the $k$-NN heat kernel $W$, the corresponding diagonal weight matrix $D$ and the Laplacian matrix $L$;
		\STATE Compute the $P$ with orthogonal columns according to Eq. (\ref{eqPXXP}) for the intermediate affinity matrix $\Delta = PP^T$;
		\STATE Get the linear projection matrix $A$ according to Eq. (\ref{eqAXLXA});
		\STATE Produce a new matrix $P$ according to Eq. (\ref{eqPXAAXP}) for the final adaptive affinity matrix $\Delta = PP^T$;
		\STATE Get linear projection $A\in\mathbb{R}^{m\times d}$ and Mahalanobis metric $M=A^TA$  by applying LPP to the affinity matrix $\Delta + D$;
	\end{algorithmic}
\end{algorithm}

Subsequently, to compute $L_\Delta$ of Eq. (\ref{eqAXLXA}) given $A$, we rewrite the affinity optimization problem with the linear projection matrix $A$ as we did in Eq. (\ref{eqXLX})
\begin{equation}
	\begin{split}		
		P = \mathop{\mathrm{arg\;min}}_{P^TP=I}\;& \Big( c+tr(A^TX^TD_\Delta XA) \\
		&+tr(A^TX^T(-PP^T)XA)\Big)
	\end{split}
	\label{eqAXDXA-PXAAXP}
\end{equation}
where we assume the final adaptive affinity matrix to be $\Delta = PP^T$ and $c=tr(A^TX^TLXA)$. The property $D_\Delta = \textbf{0}$ still holds, because of the zero mean of $XA$. Therefore, Eq. (\ref{eqAXDXA-PXAAXP}) reduces to
\begin{equation}
	P = \mathop{\mathrm{arg\;max}}_{P^TP=I}\; tr(P^TXAA^TX^TP)
	\label{eqPXAAXP}
\end{equation}

This can be solved by singular value decomposition of matrix $XA$ and taking the left-singular vectors which correspond to the largest singular values. We apply sparsification on the adaptive affinity matrix $\Delta=PP^T$ obtained from Eq. (\ref{eqPXAAXP}) and attain the sparse affinity matrix.

\begin{table*}[t]
	\caption{Clustering accuracy on image data sets(\%)}
	\label{tabAcc}
	\centering
	\begin{small}
		\begin{tabular}{|c||c|c|c|c|c|c|c|c|c|c|c|c|c|}
			\hline
			\multicolumn{1}{|l||}{} &\multicolumn{2}{|c|}{\bf AdaAM} &\multicolumn{2}{|c|}{\bf $k$-NN} &\multicolumn{2}{|c|}{\bf Cons-$k$NN} &\multicolumn{2}{|c|}{\bf DN} &\multicolumn{2}{|c|}{\bf ClustRF-Bi} & \multicolumn{2}{|c|}{\bf PCAN-$k$Means} & \multicolumn{1}{|c|}{\bf PCAN}
			\\ \hline
			& Avg & Max & Avg & Max & Avg & Max & Avg & Max & Avg & Max & Avg & Max & \\
			\hline
			UMIST    & \textbf{66.06}& \textbf{75.65}& 58.16& 65.39& 60.27& 69.22& 59.15& 66.96& 64.63& 74.44& 53.79& 56.52& 55.30\\
			COIL20   & 74.72& \textbf{87.29}& 71.89& 81.18& 75.53& 84.31& 71.95& 82.01& \textbf{76.50}& 85.07& 72.28& 83.75& 81.74\\
			USPS     & \textbf{69.36}& \textbf{69.61}& 68.25& 68.35& 68.21& 68.34& 68.08& 68.31& 58.74& 65.90& 64.04& 67.95& 64.20\\
			MNIST    & \textbf{60.84}& \textbf{61.34}& 48.13& 48.27& 47.88& 48.00& 49.72& 49.76& 51.93& 52.03& 58.93& 58.98& 59.83\\
			ExYaleB  & \textbf{54.36}& \textbf{57.87}& 24.17& 26.76& 25.63& 28.75& 24.21& 27.42& 23.10& 26.43& 25.74& 27.63& 25.89\\
			\hline
		\end{tabular}
	\end{small}
\end{table*}

Intuitively, we can iterate Eq. (\ref{eqAXLXA}) and Eq. (\ref{eqPXAAXP}) to minimize the value of objective function. However, as Fig. \ref{figitera} shows, the adaptive affinity matrix with only once iteration performs well in practice and the continuing iterations show no remarkable outperformance. 

Since the weight of nodes in the graph plays an important role in some algorithms and methods based on Normalized Cuts \cite{shi2000normalized} like LPP has the constraint relying on $D_\Delta$. In our approach we have $D_\Delta = \textbf{0}$, therefore we add the weight matrix $D$ computed from the $k$-NN heat kernel to our affinity matrix. Finally, we replace the affinity matrix in LPP with the matrix $\Delta+D$ to get the linear projection $A$ and the metric matrix $M = A^TA$.

\subsection{Sparsification Strategy}
\label{sparse}

From the optimization problem (\ref{eqPXXP}) and (\ref{eqPXAAXP}), we can observe that the matrices $XX^T$ and $XAA^TX^T$ are both low-rank matrix. Seeing that the solution of the optimization problem mentioned above is based on the singular value decomposition, this low-rank fact will result in that the column number of the  solution $P$ could be far less than the rank of $XX^T$ and $XAA^TX^T$. This process will produce a low-rank affinity matrix which leads to a progressively rank decreasing in our approach. To prevent the rank decreasing happening, we implement sparsification in our approach. The sparsification strategy may mitigate the problem of noise edges as well.

Fig. \ref{figHist} justifies our sparsification procedure by demonstrating the histogram of the magnitude of the final adaptive affinity matrix obtained from Eq.  (\ref{eqPXAAXP}) without sparsification. We can observe that most elements of the affinity matrix concentrate in the range with small magnitude, and the sparsification procedure may reserve a portion of the affinity elements which are more representative.

Inspired by the thought of $k$-NN heat kernel, we sort all the elements of affinity matrix $\Delta$ by decreasing magnitude and only reserve the first $t$ elements. We consider that the parameter $t$ is better to be in inverse proportion to the number of clusters, in which case the average elements reserved for each cluster will be proportionate to the number of data points in each cluster.  The $t$ is selected by following equation:
\begin{equation}
	t = \lfloor \frac{n^2}{\alpha c}\rfloor
\end{equation}
where $\lfloor \cdot\rfloor$ is the floor function, $n^2$ is the number of elements in $\Delta$, $c$ is the number of clusters and $\alpha$ is a coefficient.

We set  $\alpha$ to be $2.5$ for the first sparsification in the computation of the intermediate affinity matrix and set $\alpha$ to be $5$ for the second sparsification in the computation of the final adaptive affinity matrix. The $\alpha$ is decided by a rough parameter search, and it gives a stable performance in most data sets.

We summarize our algorithm in Algorithm \ref{alg1}. We set reduced dimension $m$ to be the same as the number of classes

\section{Experiments}
\label{secExp}

In this section, we conduct several experiments to demonstrate the effectiveness and efficiency of the proposed approach AdaAM. 

\subsection{Data Sets}

We evaluate the proposed approach on five 
image data sets:

{\textbf{UMIST}} The UMIST Face Database consists of 575 images of 20 individuals with 
220$\times$220 pixels  \cite{graham1998characterising}. We use the images resized to 40$\times$40 pixels 
in our experiments.

{\textbf{COIL20}} A data set consists of 1,440 images of 20 objects with discarded background  \cite{nene1996columbia}.

{\textbf{USPS}} The USPS handwritten digit database has 9,298 images of 10 digits with 16$\times$16 pixels  \cite{hull1994database}.

{\textbf{MNIST}} The MNIST database of handwritten digits has 70,000 images of 10 classes  \cite{lecun1998gradient}. In our experiments, we select the first 10,000 images of this database.

\textbf{ExYaleB} The Extended Yale Face Database B consists of 2,414 cropped images with 38 individuals and around 64 images under different illuminations per individual  \cite{KCLee05}.

The statistics of 
data sets are summarized in Tab. \ref{tabData}.
\begin{table}[ht]
	\caption{Statistics of five benchmark data sets}
	\label{tabData}
	\centering
	\begin{small}
		\begin{tabular}{c c c c}
			\hline
			Data set & \# of instances & \# of features & \# of classes\\ 
			\hline
			UMIST & 575 & 1600 & 20\\
			COIL20 & 1440 & 1024 & 20\\
			USPS & 9298 & 256 & 10\\
			MNIST & 10000 & 784 & 10\\
			ExYaleB & 2414 & 1024 & 38\\
			\hline
		\end{tabular}
	\end{small}
\end{table}

\subsection{Compared Algorithms}
We compare our approach with the other affinity learning algorithms described in Section Related Work. We adopt LPP to the affinity matrices generated by these state-of-the-art approaches to obtain the distance metric. 

\textbf{Con-$k$NN} Cons-$k$NN Consensus $k$-NNs \cite{premachandran2013consensus} with the aim of selecting robust neighborhoods.

\textbf{DN} Dominant Neighborhoods proposed in  \cite{pavan2007dominant}. 

\textbf{ClustRF-Bi} A spacial case of ClustRF-Strct \cite{zhu2014constructing}, which is also proposed in  \cite{criminisi2012decision,pei2013unsupervised}.
Due to the huge memory requirement of ClustRF-Strct on the data set with thousands instances, we implement this special case in our experiments. 

\textbf{PCAN} Projected Clustering with Adaptive Neighbors proposed in \cite{nie2014clustering}. Because PCAN is an algorithm which can generate the linear projection and clusters simultaneously, we denote the method combining the projection of PCAN and $k$-Means as PCAN-$k$Means and we also show the clustering result of PCAN in Tab. \ref{tabAcc} for reference.

We also compare our approach with the $k$-NN heat kernel affinity matrix. We use $k$-NN to denote this typical approach.

\begin{figure}[t]
	\centering
	\subfigure[UMIST]{
		\begin{minipage}{0.22\textwidth}
			\includegraphics[width=1\columnwidth]{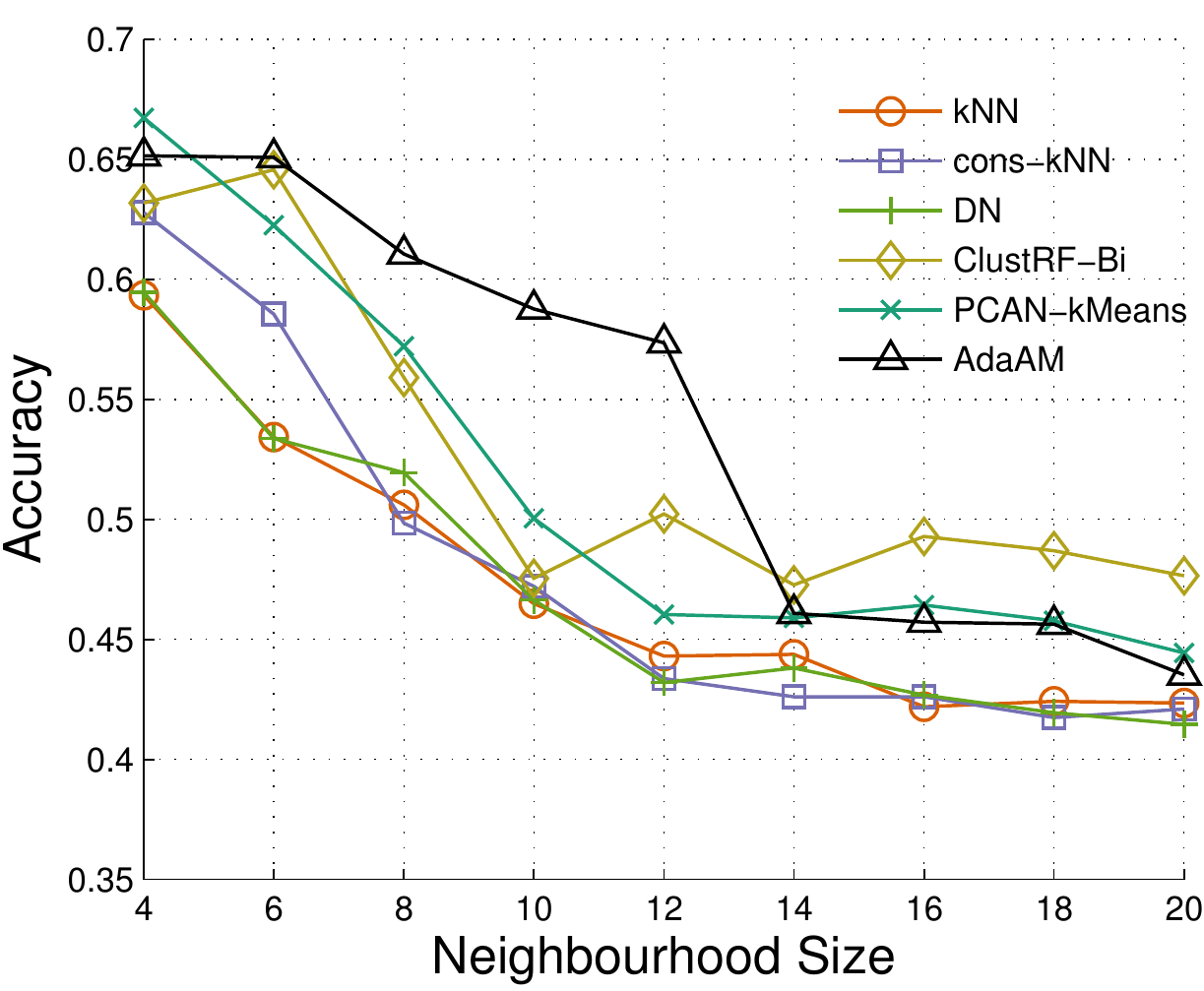}
		\end{minipage}
	}
	\subfigure[COIL20]{
		\begin{minipage}{0.22\textwidth}
			\includegraphics[width=1\columnwidth]{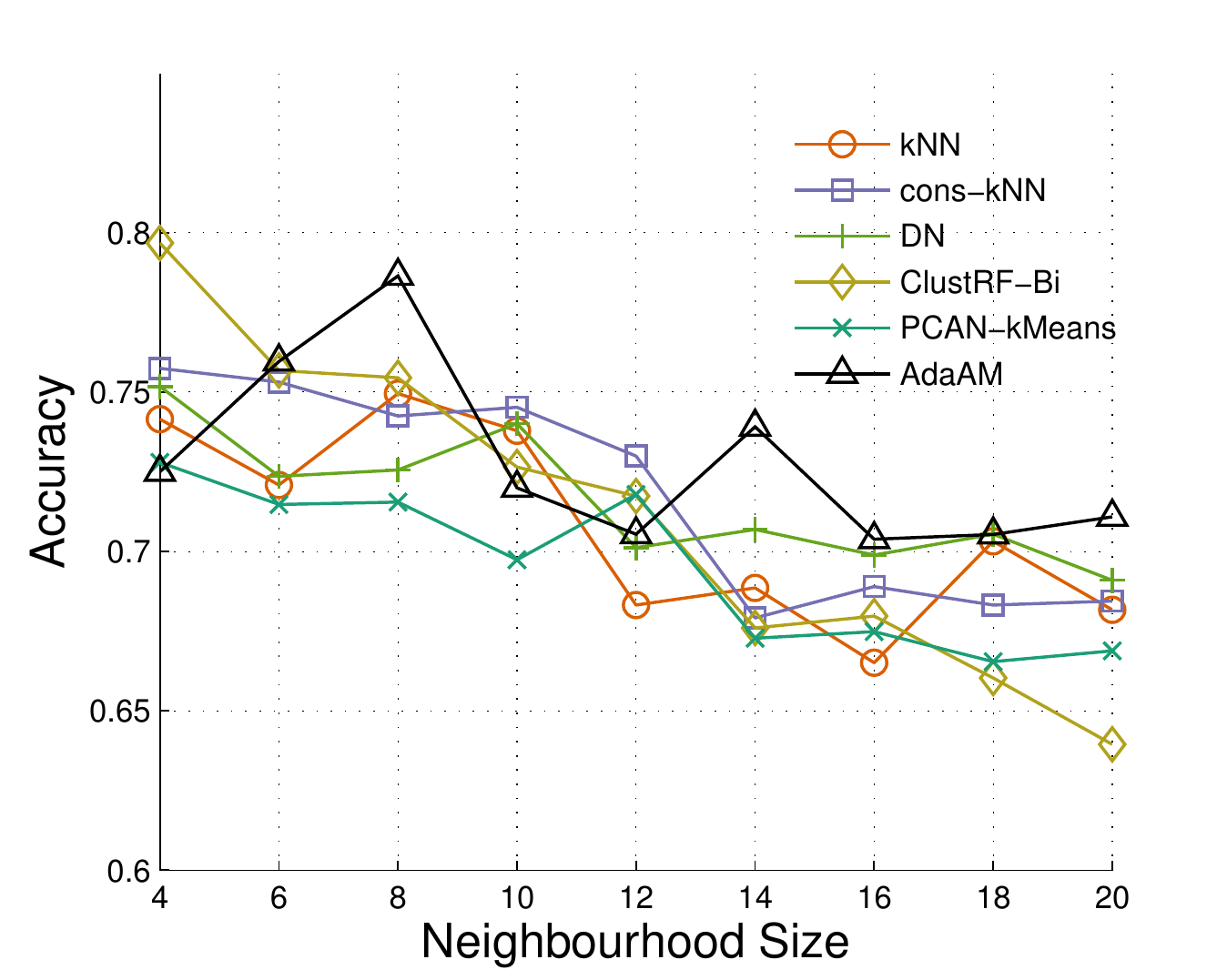}
		\end{minipage}
	}
	\subfigure[USPS]{
		\begin{minipage}{0.22\textwidth}
			\includegraphics[width=1\columnwidth]{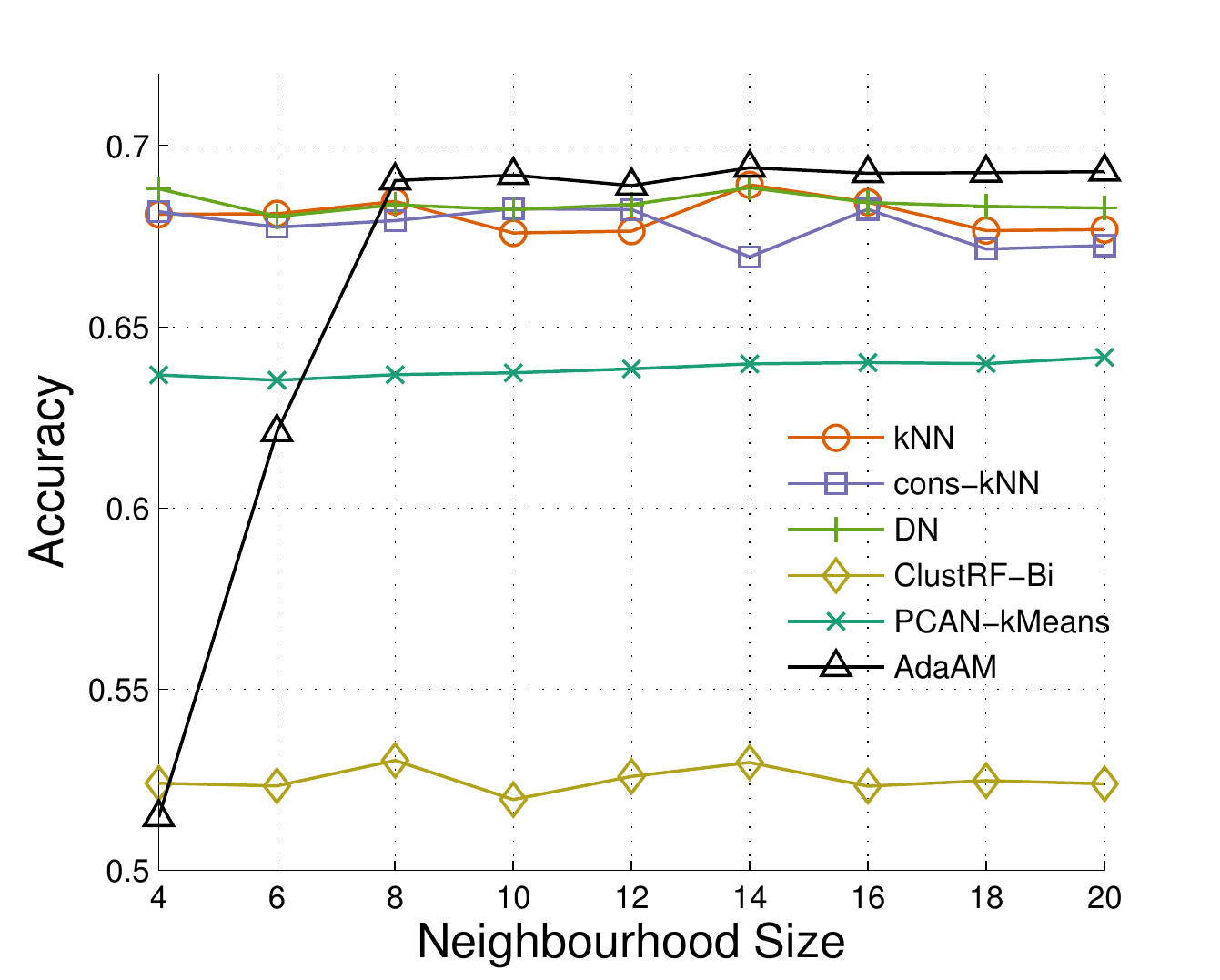}
		\end{minipage}
	}	
	\subfigure[ExYaleB]{
		\begin{minipage}{0.22\textwidth}
			\includegraphics[width=1\columnwidth]{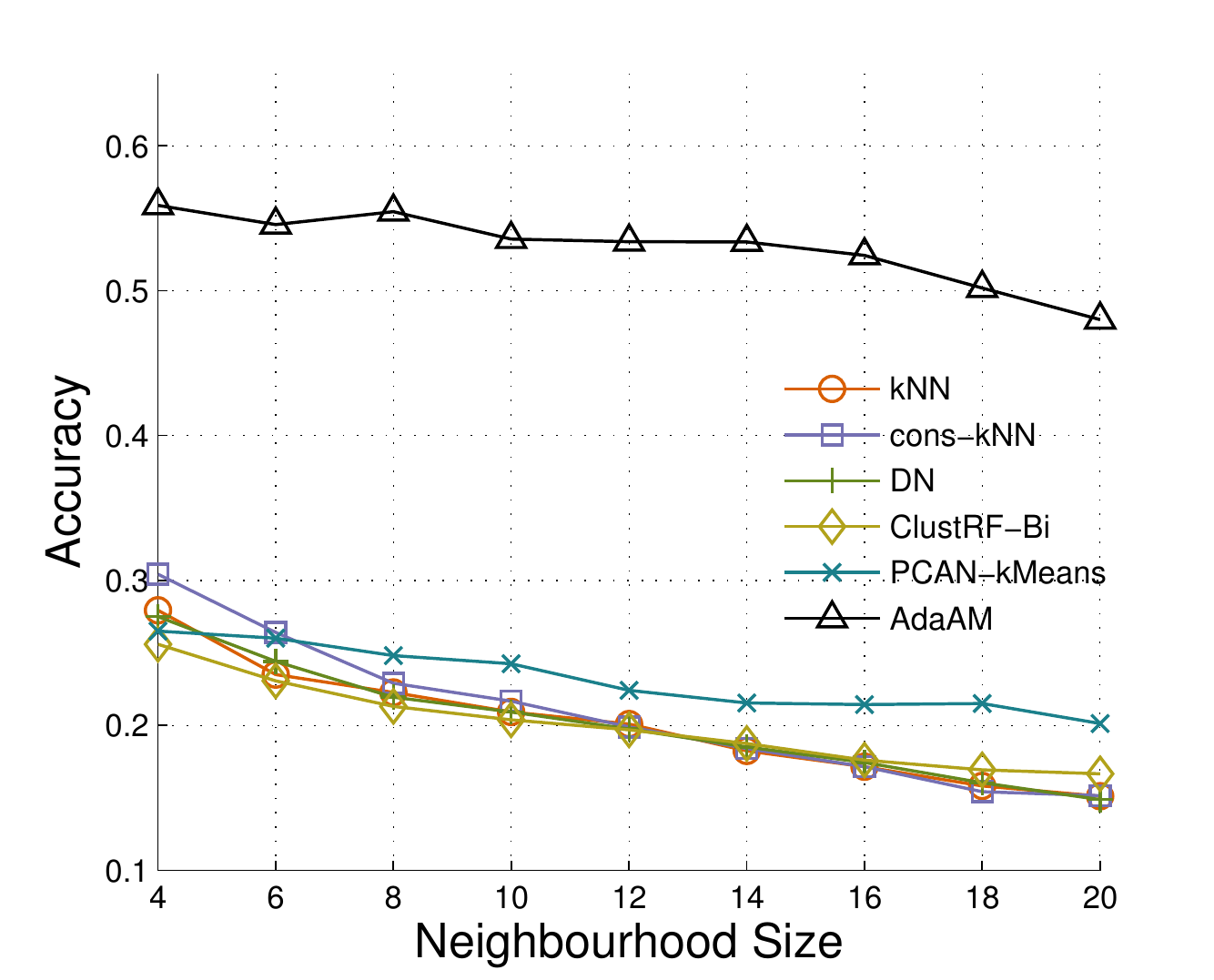}
		\end{minipage}
	}
	\caption{Comparison between different with different of neighborhood size $k$}
	\label{figSen}
\end{figure}

\subsection{Parameter Selection and Experiment Details}

Because there is no validation data set in unsupervised learning tasks, for more general case, we impose the same parameter selection criteria on all the algorithms in our experiments. We set the size of neighborhood to be $k = \mathrm{Round}(\mathrm{log}_2(n/c))$, where $n$ is the number of data instances and $c$ is the number of classes. We also set the projected dimension, which is equal to the rank of metric matrix, to be the same as the number of classes \cite{ng2002spectral}. All the other parameters in our approach are fixed in every experiment.

\begin{figure}[t]
	\centering
	\includegraphics[width=0.9\columnwidth]{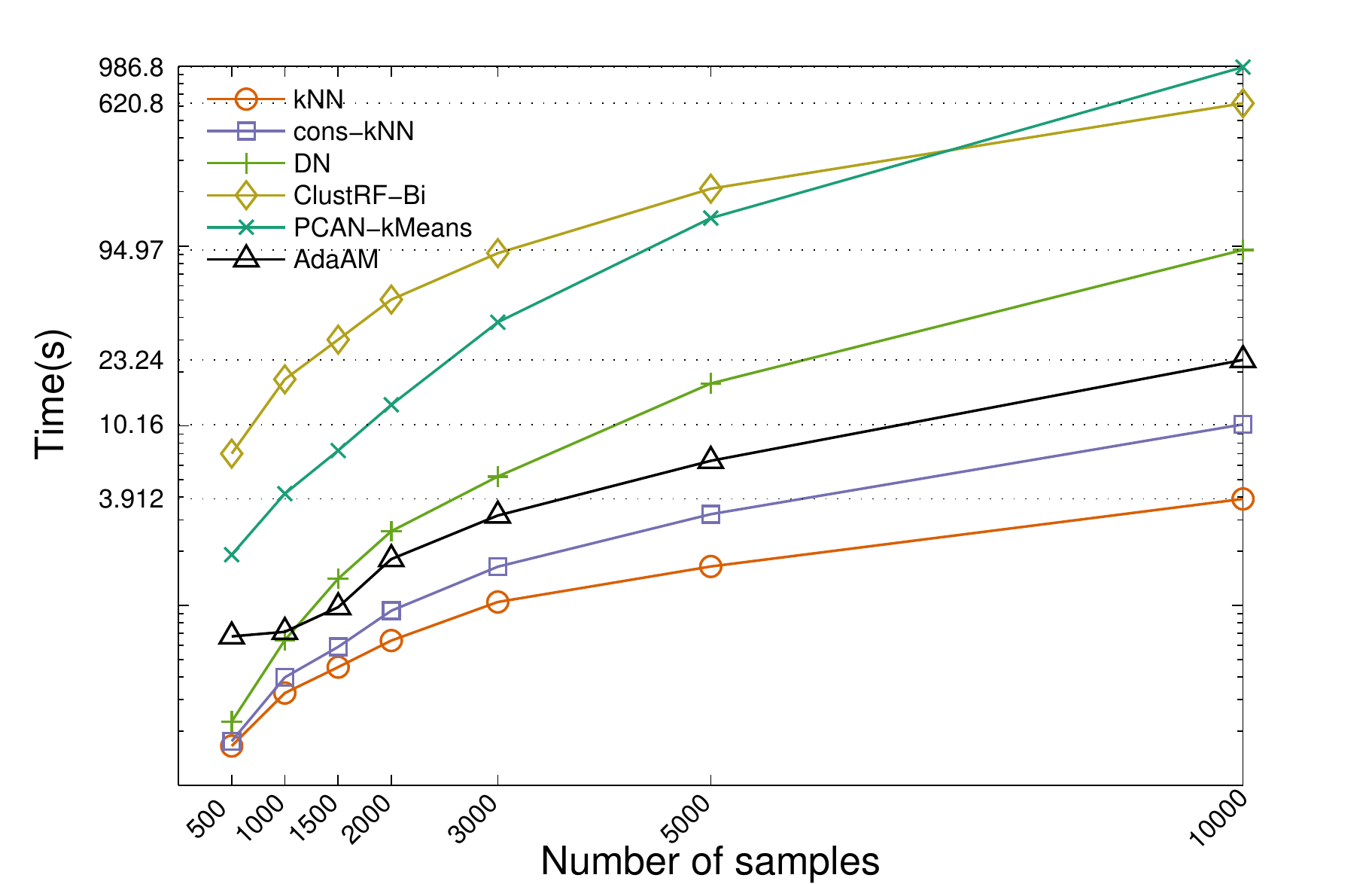}
	\caption{Time consumption of six approaches with different number of data instances}
	\label{figTime}
\end{figure}

We denote 10 times of $k$-Means as a round and select the clustering result with the minimal within-cluster sum as the result of each round of $k$-Means. We apply 100 rounds $k$-Means to each algorithms for the evaluation of the performance (Tab. \ref{tabAcc}), 10 rounds $k$-Means for the experiment of the sensitivity to the neighborhood size $k$ (Fig. \ref{figSen}) and one round $k$-Means for the experiment of execution time (Fig. \ref{figTime}).



\subsection{Experiment Results}

In the experiment of clustering accuracy, we evaluate the projection ability of AdaAM with other five algorithms on five benchmark data sets mentioned above. Tab. \ref{tabAcc} gives the average and the maximal accuracy of 100 rounds $k$-Means of each model. From Tab. \ref{tabAcc}, we can observe that superiority of AdaAM on the task of the unsupervised metric learning. In most case, AdaAM performs much better than the other approaches. Our approach attains four best results of the average accuracy and five best maximal accuracy on five data sets. We can also observe that the proposed AdaAM decisively outperforms other five methods on ExYaleB data set. Different from the other data sets, the image data in ExYaleB are properly aligned and under different illumination. This difference makes some images more similar to the image in different class under the same illumination, which result in a high rank affinity matrix. Our approach is based on a low rank approximation of the optimal affinity matrix with the ability to handle such noises in the affinity matrix. 

Since the neighborhood size $k$ selection criteria is fixed in the experiment of accuracy, which may cause the loss of the best performance, we show the trend of accuracy according to the size of neighborhood in Fig. \ref{figSen}. Fig. \ref{figSen} shows that AdaAM attains the best result in most cases and the sensitivity to  the size of neighborhood is better or comparable to the other models. Since our approach is based on the low rank approximation of the optimal affinity matrix, it requires more information from the pairwise similarity. Hence, for small $k$, baseline methods are sometimes better than our approach.

Fig. \ref{figTime} illustrates the efficiency of AdaAM by the semi-log graph of execution time with different number of data points selected from MNIST. It can be observed that our approach is a inexpensive algorithm in practice with much lower time consumption to PCAN-$k$Means, ClustRF-Bi and DN. We also show that AdaAM keeps approximately double time consumption to Cons-$k$NN with the much better performance.
\section{Conclusion}
In this paper, we present a novel affinity learning approach for unsupervised metric learning, called Adaptive Affinity Matrix (AdaAM). In our new affinity learning model, the affinity matrix is learned from the same framework of spectral clustering. More specifically, we show that the affinity learning can be reduced to a singular value decomposition problem. With the affinity matrix learned, the distance metric can be derived by some off-the-shelf approaches based on the affinity graph like LPP. Extensive experiments on clustering image data sets demonstrate the superiority of the proposed method AdaAM. 

\bibliography{bibfile1}
\bibliographystyle{IEEEbib}

\end{document}